\documentclass[letterpaper]{article} % DO NOT CHANGE THIS
\usepackage{aaai2026}  % DO NOT CHANGE THIS
\usepackage{times}  % DO NOT CHANGE THIS
\usepackage{helvet}  % DO NOT CHANGE THIS
\usepackage{courier}  % DO NOT CHANGE THIS
\usepackage[hyphens]{url}  % DO NOT CHANGE THIS
\usepackage{graphicx} % DO NOT CHANGE THIS
\usepackage{natbib}  % DO NOT CHANGE THIS AND DO NOT ADD ANY OPTIONS TO IT
\usepackage{caption} % DO NOT CHANGE THIS AND DO NOT ADD ANY OPTIONS TO IT
\usepackage{amsmath}
\usepackage{booktabs}
\usepackage{algorithm}
\usepackage{algorithmic}
\usepackage{amsfonts}

\usepackage{newfloat}
\usepackage{listings}
\DeclareCaptionStyle{ruled}{labelfont=normalfont,labelsep=colon,strut=off} % DO NOT CHANGE THIS
\floatstyle{ruled}
\newfloat{listing}{tb}{lst}{}
\floatname{listing}{Listing}
\title{MINGLE: VLMs for Semantically Complex Region Detection in Urban Scenes}
\author {
    Liu Liu\textsuperscript{\rm 1}\equalcontrib ,
    Alexandra Schild\textsuperscript{\rm 2}\equalcontrib ,
    Marco Cipriano\textsuperscript{\rm 2}, 
    Fatimeh Al Ghannam\textsuperscript{\rm 1}, 
    Freya Tan\textsuperscript{\rm 1},\\ 
    Gerard de Melo\textsuperscript{\rm 2},
    Andres Sevtsuk\textsuperscript{\rm 1}
}
\affiliations {
    \textsuperscript{\rm 1}Massachusetts Institute of Technology\\
    \textsuperscript{\rm 2}Hasso Plattner Institute\\
    lyons66@mit.edu, aleksandra.kudaeva@hpi.de
}

\begin{document}
\maketitle

\begin{abstract}

Understanding group-level social interactions in public spaces is crucial for urban planning, informing the design of socially vibrant and inclusive environments. Detecting such interactions from images involves interpreting subtle visual cues such as relations, proximity, and co-movement—semantically complex signals that go beyond traditional object detection. To address this challenge, we introduce a social group region detection task, which requires inferring and spatially grounding visual regions defined by abstract interpersonal relations. We propose MINGLE (Modeling INterpersonal Group-Level Engagement), a modular three-stage pipeline that integrates: (1) off-the-shelf human detection and depth estimation, (2) VLM-based reasoning to classify pairwise social affiliation, and (3) a lightweight spatial aggregation algorithm to localize socially connected groups. To support this task and encourage future research, we present a new dataset of 100K urban street-view images annotated with bounding boxes and labels for both individuals and socially interacting groups. The annotations combine human-created labels and outputs from the MINGLE pipeline, ensuring semantic richness and broad coverage of real-world scenarios.

\end{abstract}

\begin{links}
      \link{Code and Dataset}{https://github.com/lyons66/MINGLE}
 \end{links}

\section{Introduction}

Object detection is a foundational task in computer vision, and state-of-the-art models, such as YOLO and Grounding DINO, excel at localizing discrete, well-defined objects, typically described by simple nouns or phrases \cite{liu2024grounding, ren2016faster, Jocher_Ultralytics_YOLO_2023, cheng2024yolo}. However, an increasing number of real-world applications require the detection of regions defined by semantically complex or abstract criteria, e.g., interactions of objects, social behaviors, or environmental context. Such tasks require a deeper understanding of context, relationships, and semantics than conventional object detection models are equipped to handle.

Vision-language models (VLMs), trained on massive image–text datasets, possess strong contextual reasoning and semantic abstraction capabilities. Yet, they are generally incapable of accurately grounding rich visual descriptions to specific regions in an image. This limitation arises from two main factors: (1) sequential training objectives tailored to text, and (2) the lack of datasets and data creation methods for semantically complex region grounding during training.

\begin{figure}[ht]
\centering
\includegraphics[width=0.9\columnwidth]{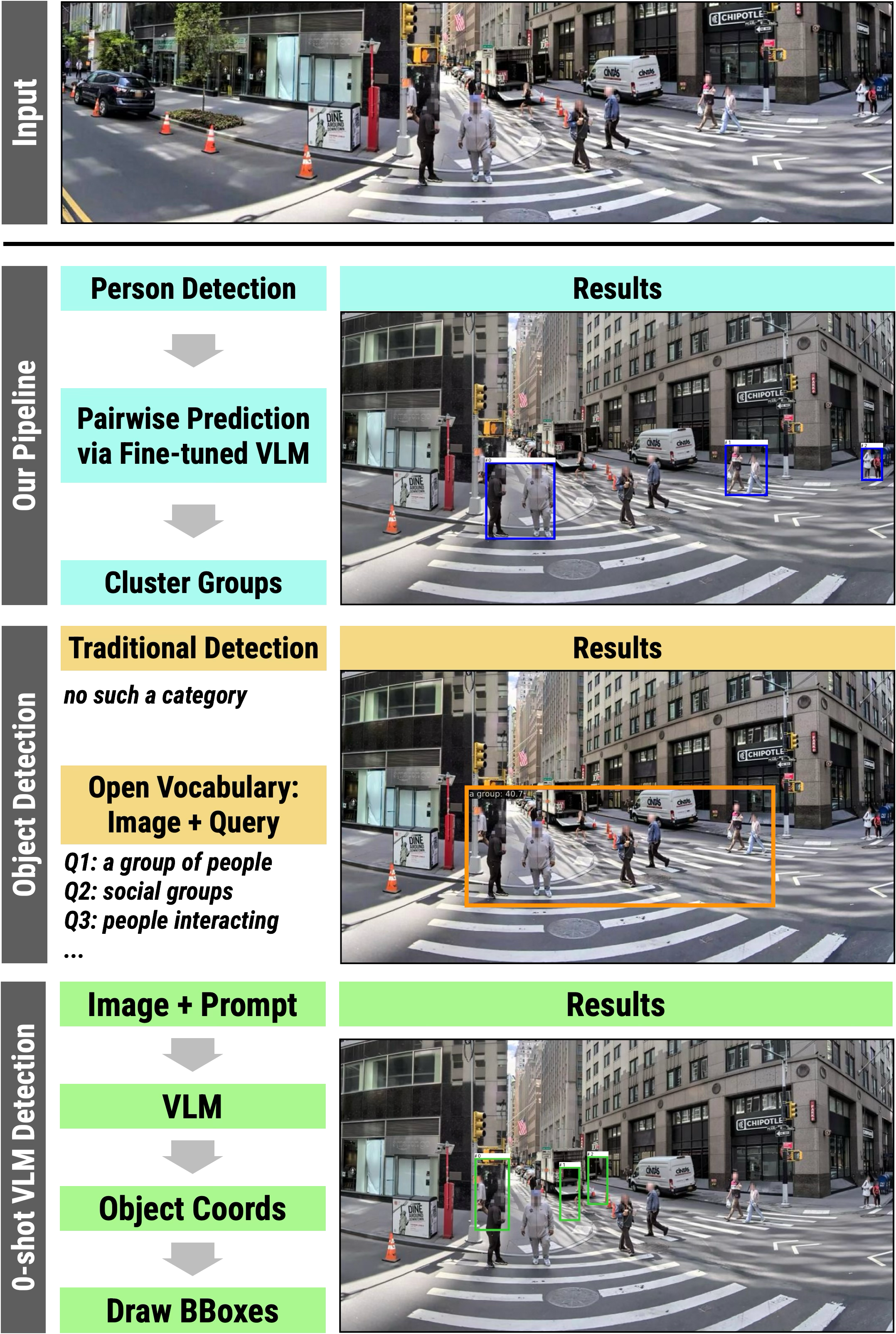}
\caption{Comparison of our method with zero-shot object detection and standard VLM-based methods. Our pipeline enables localized detection of socially interacting human groups in complex urban scenes.}
\label{fig1}
\end{figure}

In this paper, we propose the task of social group region detection in urban scenes—detecting and localizing groups of individuals who are socially interacting. Unlike traditional object detection, which targets physical entities, this task requires identifying abstract, emergent visual patterns involving multiple people. These include diverse visual cues such as co-movement, proximity, body posture, and spatial formation—signals that define social groups rather than isolated entities.

From an urban design perspective, detecting social groups in public spaces has long been a key focus in pedestrian observational studies. Seminal works by Whyte \cite{whyte1980social}, Gehl \cite{gehl2011life}, and Jacobs \cite{jacobs_death_1961, jacobs1985looking} emphasize that observing how people gather, linger, and interact in cities is crucial to understanding urban vitality. These studies go far beyond simply counting individuals: Inferring social groups is central to evaluating how built environments support or inhibit human interaction. Such knowledge informs the creation of healthier, more inclusive, and socially vibrant urban environments.

Still, research in this area remains constrained by the lack of scalable, semantically rich annotations. Existing datasets typically either lack sufficient scale, fail to capture complex interaction cues, or require labor-intensive manual labeling that does not scale to a city- or country-level analysis.

To address this gap, we propose MINGLE, a modular three-stage pipeline for detecting socially interacting human groups in street-view imagery. MINGLE leverages both existing state-of-the-art object detection and depth estimation models as well as the semantic comprehension and reasoning capabilities of VLMs. First, we apply the best available off-the-shelf object detection and depth estimation models to detect all humans in the scene and create an image depth map. Second, using a novel human-annotated dataset, we fine-tune a vision-language model to classify whether a pair of individuals belong to the same social group. Finally, we apply an efficient pair-aggregation algorithm to identify social group regions.

Using MINGLE, we construct a new dataset of 100K images with annotations for both individual people and social groups, making it the first large-scale dataset focused on semantically complex social interaction regions in urban outdoor scenes, not limited to a particular interaction type. Our labels combine high-quality manual annotations and carefully validated pseudo-labels propagated through our pipeline.

This dataset serves as a valuable benchmark and resource for research in relational vision, social perception, and urban computing. It can be used to fine-tune VLMs for grounding relational concepts, to evaluate multi-object interaction models, or to explore high-level visual understanding.

Our contributions are threefold:
\begin{itemize}
\item We introduce a new vision-language task: social group region detection, which involves grounding relational and semantic concepts in visual space;
\item We propose a scalable pipeline that combines object detection, depth estimation, and VLM reasoning to localize social interaction regions;
\item We release a large-scale dataset of urban scenes with bounding boxes for both individual people and socially interacting groups, enabling future work in this domain.
\end{itemize}

\noindent Beyond urban planning, social group region detection has broad applications in social dynamics analysis, autonomous navigation, AR/VR, robotics, as well as potentially harmful applications such as surveillance. We hope that this work encourages the vision community to explore more nuanced and relational forms of scene understanding—moving beyond object detection toward semantically complex scene comprehension and region detection.

\section{Related Work}

\subsection{Social Interaction Research}

Detecting social groups has long been a key focus in pedestrian observational studies. Starting from classic works by Whyte, Gehl, and Jacobs, more recent studies in urban planning revisit criteria of public space liveliness and emphasize that social interactions play a central role \citep{whyte1980social, jacobs_death_1961, jacobs1985looking, gehl2011life, mehta2021revisiting, qi2024understanding}. 

However, conducting large-scale analyses of group-level social behavior has been limited by the labor- and cost-intensive nature of manual fieldwork. Advances in AI and computer vision aim to address this challenge, offering tools to automate the detection and analysis of social dynamics at scale. Despite significant progress in human–human interaction detection, most computer vision approaches focus on relatively constrained scenarios. With existing street-view benchmarks being limited in scale \cite{hosseini2024elsaevaluatinglocalizationsocial}, large-scale datasets such as JRDB \cite{ehsanpour2022jrdb, jahangard2024jrdb} and AVA \cite{gu2018ava} primarily support action or gesture recognition in surveillance or indoor contexts. Others target pairwise or small-group social signal understanding, often framed as classification problems based on a closed set of body pose, gaze, or proximity cues \cite{wang2023humantohumaninteractiondetection, wei2024nonverbal}. Although these datasets and models offer valuable insights, they do not generalize to open-world, urban street environments, where social formations are more ambiguous and embedded in complex spatial and visual contexts.

\subsection{Open-Vocabulary Object Detection}

\begin{figure}[t] % this can be updated
\centering
\includegraphics[width=1\columnwidth]{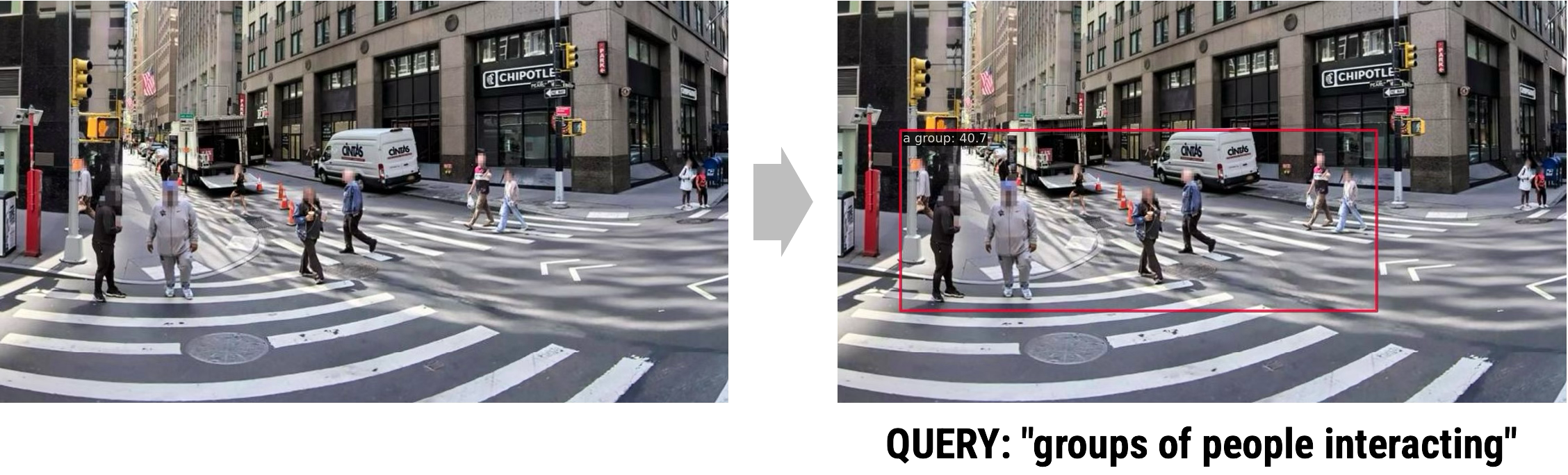}
\caption{The result of OVD for social group detection.}
\label{fig_ovd}
\end{figure}

Language-based object detection expands the fixed categories of traditional closed-set detection to free-form descriptions. Current commonly used open-vocabulary object detection and segmentation models, such as YOLO-World, Grounding-DINO, and Grounded SAM, have demonstrated reliable performance on standard object detection or segmentation benchmarks \cite{cheng2024yolo, liu2024grounding, ren2024grounded}. However, reliance on BERT-based language models limits their ability to comprehend semantically complex language structures. As a result, OVD models struggle to interpret abstract natural language instructions for precise visual grounding, particularly when handling complex instructions and visual scenarios. In particular, for social group region detection, they often fail to interpret and reason between complex inter-group relationships and the surrounding urban context  (Figure~\ref{fig_ovd}). 

\subsection{VLMs for Localization, Object Detection, and Segmentation}

Large language models (LLMs) and vision-language models (VLMs) excel at understanding semantically complex commands and visual contexts. However, most general-purpose VLMs lack built-in support for precise spatial tasks such as segmentation and object detection, mostly due to the lack of such training data and specialized training procedures. Recent studies confirm that a Transformer architecture trained with a sufficiently large amount of localization data can achieve superior performance on object detection and segmentation tasks \cite{xiao2023florence2advancingunifiedrepresentation}.

The literature highlights several key limitations of current VLMs in region-level localization tasks. Most generalist VLMs output text-based responses without explicit or reliable spatial grounding, limiting their direct applicability to detection or segmentation. Specialized adaptations have emerged, including learnable coordinate tokens and decoupled localization pipelines, e.g., ROD-MLLM \cite{yin2025rod}, designed to better align textual instructions with precise visual regions.

Moreover, models such as LISA \cite{lai2024lisareasoningsegmentationlarge} and \mbox{u-LLaVA} \cite{xu2024ullavaunifyingmultimodaltasks} introduce segmentation tokens as prompts within VLMs to improve mask prediction, yet these often still focus on single-object instances, lacking robust mechanisms for multi-object or group-level reasoning. Subsequent approaches such as GLaMM \cite{rasheed2024glammpixelgroundinglarge}, PerceptionGPT \cite{pi2023perceptiongpteffectivelyfusingvisual}, and PixelLM \cite{ren2024pixellmpixelreasoninglarge} extend this with condition tokens for multi-object grounding, but their effectiveness degrades with increasing instruction complexity and relational reasoning demands.

\subsection{Semantically Complex Object Detection and Segmentation}

\begin{figure*}[!t]
\centering
\includegraphics[width=1\textwidth]{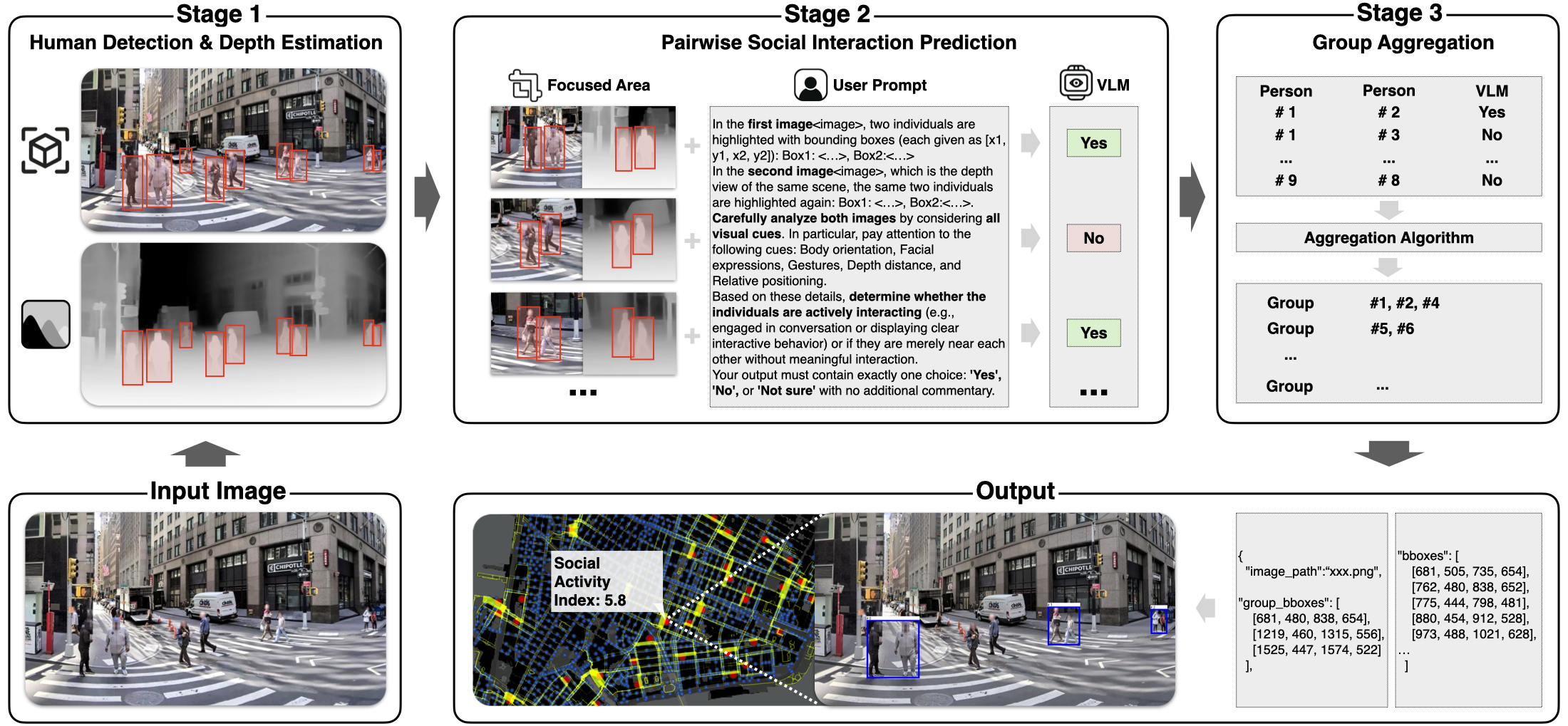}
\caption{Illustration of the three-stage pipeline for detecting semantically complex social interaction regions.}
\label{fig:method}
\end{figure*}

A recent review of VLM usage for object detection and segmentation underscores that despite progress achieved through multi-modal pretraining, semantically complex region detection—especially for abstract or relational concepts—remains an open challenge \cite{feng2025visionlanguagemodelobjectdetection}. Their analysis reveals persistent difficulties in grounding abstract queries, rejecting false positives, and accurately localizing regions defined by high-level semantics rather than explicit category labels.

To address this, recent work such as Ground-V \cite{zong2025groundvteachingvlmsground} has taken a step forward by explicitly training VLMs to align natural language descriptions—including complex and compositional instructions—with pixel-level regions. Ground-V demonstrates improved grounding for instructions involving actions and scene context, but it primarily targets synthetic or simple compositional examples and lacks evaluation in socially grounded, real-world scenarios.

As existing methods fail when regions are described using semantically rich concepts, there is a clear need for methods that extend VLMs toward reasoning-driven localization in complex real-world scenes. Our work fills the gap by focusing on social interaction regions in urban scenes—an example of semantically complex region detection with real-world societal impact. By combining pairwise reasoning with spatial aggregation and introducing the MINGLE dataset, we contribute toward enabling robust detection of abstract, relational visual concepts that current models struggle to handle, as well as to urban planning analysis, enabling the design of healthier, more inclusive cities.

\section{Method}

In this work, we introduce a new task—social group region detection, which requires generating a bounding box given a  query text involving implicit reasoning capabilities. The main difference to the conventional object detection task is that the query text is not limited to a straightforward object reference (e.g., “the person”), but is a more complex description requiring complex reasoning and world knowledge, for example:

\begin{quote}
Based on visual cues, determine whether the individuals are actively interacting (e.g., engaged in conversation or displaying clear interactive behavior) or if they are merely near each other without meaningful interaction.
\end{quote}

\noindent This task is characterized by three key challenges: 1) the complexity of the text query and visual context; 2) multiple regions that satisfy criteria simultaneously, and 3) the potential for hallucinated references of regions not present in the scene. In the following section, we describe a three-stage workflow that enables overcoming these challenges to efficiently detect socially affiliated groups of people in street-view images (see Figure \ref{fig:method}). 

\subsection{Social Group Region Detection Task}

We formulate the problem as follows. Let
\[
\mathcal{I} = \{ I_1, I_2, \ldots, I_M \mid I_m \in \mathbb{R}^{H \times W \times 3} \}
\]
be a set of RGB streetview images. For each image $I_m \in \mathcal{I}$, let
\begin{align*}
\mathcal{P}^{m} &= \{ p^{(m)}_1, p^{(m)}_2, \ldots, p^{(m)}_K \mid K \geq 0 \},\\
\mathcal{G}^{m} &= \{ g^{(m)}_1, g^{(m)}_2, \ldots, g^{(m)}_L \mid L \geq 0 \}
\end{align*}
denote, respectively, the set of detected single person bounding boxes and socially affiliated group bounding boxes in image $I_m$, where each $g^{m}_i$ for image $I_m$ contains at least two individual detections from $\mathcal{G}^m$ such that $\{p^{m}_j \in \mathcal{P}^{m} \mid p^{m}_j \subseteq g^{m}_i \}| \geq 2$.

The goal is to detect all such regions containing social groups that consist of two or more socially connected individuals in each image $I_m$. 

Let $\mathcal{R}: \mathbb{R}^{H \times W \times 3} \rightarrow \mathcal{G}$ be the full pipeline that maps an image $I_m$ to a set of socially affiliated group region bounding boxes $\mathcal{G}$.

In order to ensure that pipeline results do not suffer from hallucinated references or struggle with multi-region detections, as is common for VLM-based object detection methods \cite{zong2025groundvteachingvlmsground}, we propose a three-stage pipeline leveraging existing object-detection models' capabilities.

\subsection{Three-Stage Pipeline}
The proposed pipeline combines off-the-shelf human detection, VLM-based pairwise reasoning, and a lightweight spatial aggregation algorithm to detect socially affiliated groups in street-view images (see Figure~\ref{fig:method}).

\noindent\textbf{Stage 1: Single Person Detection} 

\noindent In order to address two common critical issues faced by VLMs applied to object detection—hallucinated references and multiple detections, we design our pipeline based on a robust, high-precision, and recall human detector. 

Given an image $I_m \in \mathcal{I}$, we first apply a pre-trained human detector
\[
\mathcal{D}: \mathbb{R}^{H \times W \times 3} \rightarrow \mathcal{P}
\]
to predict a set of bounding boxes for the \texttt{person} category, where each \( p_i \) is an axis-aligned bounding box \( b_i \) with an associated confidence score \( s_i \in [0,1] \). We retain only high-certainty predictions by discarding detections with \(s^{(m)}_i < \tau_{\text{det}}\), where $\tau_{\text{det}}$ can be empirically chosen based on the human-predictor model and the desired precision–recall trade-off. 

We instantiate $\mathcal{D}$ with the ATSS-Swin-L-DyHead model~\cite{zhang2020bridging}. However, Stage 1 is detector-agnostic: \( \mathcal{D} \) may be replaced by any modern human detector, e.g., a YOLOv8 variant \cite{Jocher_Ultralytics_YOLO_2023}, for further improvements without requiring any modifications to the downstream pipeline.

\bigskip
\noindent\textbf{Stage 2: Pairwise Social-Interaction Prediction}

\noindent For every unordered pair of bounding boxes \((p^{(m)}_i , p^{(m)}_j)
\), we fine-tune Qwen2-VL-7B and Qwen2.5-VL-7B to perform classification (“Yes”, “No”, or “Not sure”), indicating whether the pair belongs to the same social group using a human-annotated dataset (see Dataset Construction section). 

\paragraph{Visual inputs.}  
We first form a padded union box  
\[
U^{(m)}_{ik}= \text{Pad}\!\bigl(\text{BBoxUnion}(b^{(m)}_i,b^{(m)}_k)\bigr),
\]
then crop the identical region from (i) the RGB image \(I_m\) and (ii) the predicted depth map \(D_m\) obtained with EVP \cite{lavreniuk2024evp}.  
These two crops,
\[
F^{(m)}_{ik,\mathrm{rgb}}=\text{Crop}(I_m,U^{(m)}_{ik}),
\]
\[
F^{(m)}_{ik,\mathrm{depth}}=\text{Crop}(D_m,U^{(m)}_{ik})
\]
are both overlaid with 1-pixel bounding boxes for the two individuals, so the model receives precisely aligned RGB–depth views focused on the candidate pair.

\paragraph{Depth-aware textual cues.}  
Within each original bounding box, we take all depth pixels in \(D_m\) and compute their medians, \(d^{(m)}_i\) and \(d^{(m)}_k \in[0,255]\).
These two scalars, together with their absolute difference \(|d^{(m)}_i-d^{(m)}_k|\), are inserted verbatim into the prompt text (see Figure \ref{fig:method}).

Thus, the model simultaneously accounts for (i) fine-grained RGB appearance, (ii) relative geometry in the depth crop, and (iii) explicit numeric depth values, enabling robust reasoning even when visually adjacent persons are separated in 3-D space.

In order to eliminate highly unlikely pairs and make the inference more scalable, we propose an algorithm that allows reducing the number of VLM calls needed for reliable group detection (see Algorithm \ref{alg:pairwise_vlm}).

\begin{algorithm}[ht]
\caption{Distance and Depth Filtering Algorithm}
\small
\label{alg:pairwise_vlm}
\textbf{Input}: List of bounding boxes $\mathcal{P} = \{p_1, p_2, \ldots, p_N\}$ with depth values $z_i$ \\
\textbf{Parameters}: Distance threshold $\tau_d$, depth threshold $\tau_z$ \\
\textbf{Output}: Pairwise relation matrix $\mathcal{R} \in \{ \texttt{Yes}, \texttt{No}, \texttt{Not~Sure} \}^{N \times N}$

\begin{algorithmic}[1]
\STATE Initialize $\mathcal{R}_{i,j} \leftarrow \texttt{Not~Sure}$ for all $i,j$
\FOR{each $p_i \in \mathcal{P}$}
    \FOR{each $p_j \in \mathcal{P}$}
        \IF{$i = j$}
            \STATE \textbf{continue}
        \ENDIF
        \IF{$d(p_i, p_j) > \tau_d$}
            \STATE $\mathcal{R}_{i,j} \leftarrow \texttt{No}$
            \STATE \textbf{continue}
        \ENDIF
        \IF{$|z_i - z_j| > \tau_z$}
            \STATE $\mathcal{R}_{i,j} \leftarrow \texttt{No}$
            \STATE \textbf{continue}
        \ENDIF
        \STATE result $\leftarrow$ VLM$(p_i, p_j)$ \COMMENT{Returns Yes/No/Not Sure}
        \STATE $\mathcal{R}_{i,j} \leftarrow$ result
    \ENDFOR
\ENDFOR
\STATE \textbf{return} $\mathcal{R}$
\end{algorithmic}
\end{algorithm}

We choose thresholds $\tau_z$ and $\tau_d$ empirically based on the trade-off between computational speed vs.\ group detection accuracy (see Results section).

\noindent\textbf{Stage 3: Group Aggregation and Region Generation}

\noindent Given the answers from Stage 2 for all unordered person pairs \((p^{(m)}_i, p^{(m)}_k)\), we construct the social groups by applying a greedy clustering algorithm. 

Let \(J^{(m)}: \mathcal{P}_{I_m} \times \mathcal{P}_{I_m} \rightarrow \{ \text{Yes}, \text{No}, \text{Not sure}\}\) be the set of pairwise judgments indicating whether two individuals belong to the same group.

We initialize each person as a singleton cluster, then iteratively merge the pair of clusters \((C_a, C_b)\) that maximally increases an agreement score. The score is defined based on whether merging the clusters aligns with the pairwise annotations in \(J^{(m)}\). If \((p^{(m)}_i, p^{(m)}_k) = \text{"Yes"}\), then placing \(i\) and \(k\) in the same cluster increases the score, while if the answer is \text{"No"} or \text{"Not sure"}, it does not.

This process continues greedily until no further improvement can be made. We define valid social groups as those clusters with cardinality at least 2, i.e., \(|g^{(m)}_j| \ge 2\).

For each group $g^{(m)}_j$, we compute a group bounding box $B^{(m)}_j$ by taking the smallest and largest coordinates from all individual person boxes in the group:
\[
B^{(m)}_j = \left(\min x^{(1)}_p, \min y^{(1)}_p, \max x^{(2)}_p, \max y^{(2)}_p\right), p \in g^{(m)}_j
\]
where each person's box is $b_p = (x^{(1)}_p, y^{(1)}_p, x^{(2)}_p, y^{(2)}_p)$. These group boxes provide a compact spatial summary of social groups for visualization and evaluation.

\section{Experiments}
\subsection{Fine-Tuning Dataset Construction}\label{pairdata}
To construct the fine-tuning dataset, we collected 79,265 human-labeled annotations for pairwise social interactions between single people in the images. In each training image, exactly two individuals are highlighted with bounding boxes. Human annotators were instructed to identify whether the indicated individuals belong to the same social group (``yes''), are independent from each other (``no''), or it is not possible to determine (``not sure''). To ensure that the resulting dataset is balanced with regard to ``yes'' and ``no'' answers, the annotation process consisted of two iterations: (1) Human annotators received 46K randomly selected street-view images. (2) Based on the analysis of the first iteration (skewed towards ``no'' answers), we pre-selected pairs with individuals potentially belonging to the same group based on the first-stage fine-tuned Qwen-2 (7B) model answers. The resulting dataset distribution is described in Table \ref{tab:pairdataset}.

\begin{table}[h]
\centering
\begin{tabular}{lrrrr}
    \textbf{Split} & \textbf{Pairs} & \textbf{"Yes"} & \textbf{"No"} & \textbf{"Not Sure"} \\
    \midrule
    Overall & 79,265 & 29,716 & 45,029 & 4,520 \\
    Train & 76,265 & 28,341 & 43,654 & 4,270 \\
    Test & 2,000 & 925 & 925 & 150 \\
    Validation & 1,000 & 450 & 450 & 100 \\
\end{tabular}
\caption{Pairwise Relationships Dataset Overview.}
\label{tab:pairdataset}
\end{table}

For the model training, the pairwise-relationships dataset is additionally split into different subsets. Variants differ by dataset sampling (balanced vs.\ unbalanced) and pre-processing (deletion or replacement of low-confidence pairs, i.e., those labeled as ``not sure'', where ``YN'' indicates that all ``not sure'' answers were omitted).

\begin{table*}[t]
\centering
\begin{tabular}{lllc
                *{2}{c}  % F1 (Yes, No)
                *{2}{c}  % Precision (Yes, No)
                *{2}{c}  % Recall (Yes, No)
                }
\toprule
\textbf{Model} & \textbf{Model Size} & \textbf{Fine-tuning Method} & \textbf{Fine-tuning Data}
& \multicolumn{2}{c}{$\boldsymbol{F_1}$} 
& \multicolumn{2}{c}{\textbf{Precision}} 
& \multicolumn{2}{c}{\textbf{Recall}} \\
\cmidrule(lr){5-6} \cmidrule(lr){7-8} \cmidrule(lr){9-10}
& & & & \textbf{Yes} & \textbf{No} & \textbf{Yes} & \textbf{No} & \textbf{Yes} & \textbf{No} \\
\midrule
\multicolumn{9}{l}{\textbf{Evaluation result of first round fine-tuning}} \\
\midrule
Qwen2-VL & 7B & LoRA (rank 64) & 29k-YN & 0.64 & 0.57 & 0.56 & 0.69 & 0.75 & 0.49 \\
Qwen2.5-VL & 7B & LoRA (rank 64) & 29k-YN & 0.65 & 0.53 & 0.55 & 0.72 & 0.82 & 0.42 \\
Qwen2.5-VL & 7B & LoRA (rank 128) & 29k-YN & 0.66 & 0.39 & 0.52 & 0.78 & 0.91 & 0.26 \\

\midrule
\multicolumn{9}{l}{\textbf{Evaluation result of second round fine-tuning}} \\
\midrule
Qwen2-VL & 7B & LoRA (rank 64) & 61k-Balanced & 0.68 & 0.62 & 0.59 & 0.75 & 0.80 & 0.53 \\
Qwen2-VL & 7B & LoRA (rank 64) & 76k-Unbalanced & 0.70 & 0.72 & 0.67 & 0.75 & 0.73 & 0.69 \\
Qwen2-VL & 7B & LoRA (rank 64) & 61k-Balanced-YN & 0.72 & 0.70 & 0.65 & 0.78 & 0.79 & 0.63 \\
Qwen2-VL & 7B & LoRA (rank 64) & 76k-Unbalanced-YN & 0.68 & 0.65 & 0.61 & 0.75 & 0.78 & 0.57 \\
\multicolumn{9}{l}{\textbf{Baseline results (no fine-tuning)}} \\
\midrule
Phi 3.5  & 4.2B & --- & --- & 0.57 & 0.24 & 0.45 & 0.47 & 0.79 & 0.16 \\
Qwen2-VL & 7B & --- & --- & 0.53 & 0.54 & 0.50 & 0.58 & 0.58 & 0.50 \\
Qwen2.5-VL & 7B & --- & --- & 0.10 & 0.69 & 0.57 & 0.54 & 0.05 & 0.97 \\
Qwen2-VL & 72B & --- & --- & 0.62 & 0.58 & 0.56 & 0.67 & 0.71 & 0.51 \\
Qwen2.5-VL & 72B & --- & --- & 0.59 & 0.56 & 0.54 & 0.64 & 0.68 & 0.49 \\
Pixtral & 12B & --- & --- & 0.52 & 0.51 & 0.48 & 0.56 & 0.46 & 0.57 \\
Gemma 3 & 24B & --- & --- & 0.27 & 0.67 & 0.49 & 0.54 & 0.18 & 0.84 \\
LLama3.2-Vision & 11B & --- & --- & 0.46 & 0.30 & 0.40 & 0.38 & 0.55 & 0.25 \\
GPT 4o & --- & --- & --- & 0.00 & 0.70 & 0.00 & 0.54 & 0.00 & 1.00 \\
Claude-Sonnet& --- & --- & --- & 0.24 & 0.66 & 0.47 & 0.54 & 0.16 & 0.85 \\
\end{tabular}
\caption{Model Performance on pairwise relationship classification task. Fine-tuning results in the upper part of the table, zero-shot VLM performance in the lower part.}
\label{tab:fine_tuning_results}
\end{table*}

\subsection{Model Selection}
To identify a strong yet efficient model for pairwise social affiliation classification, we benchmarked several open-source and commercial VLMs on a curated set of 2,000 manually validated image pairs. As shown in Table~\ref{tab:fine_tuning_results}, larger models like Qwen2-VL (72B) outperform others in overall $F_1$. However, smaller variants—especially Qwen2-VL (7B)—obtain just marginally lower effectiveness, while offering greater robustness to overfitting, lower memory consumption, and significantly faster inference. The latter is crucial for urban research,\footnote{Urban analysis requires working with vast amounts of data; e.g., for New York City alone, the number of street views exceeds 8M.} and therefore Qwen2-VL (7B) and Qwen2.5-VL (7B) were selected for further fine-tuning.

\subsection{Prompt Generation} 
To create suitable text prompts for guiding the model towards social group understanding, we evaluated a variety of formulations augmented with additional visual and textual cues (e.g., bounding box coordinates for individual people, numeric depth and distance values, etc.). We determined that the best results are achieved when the model receives inputs and prompts as illustrated in Figure \ref{fig:method}. We release the nine best-performing prompt versions in the project repository.

\subsection{Evaluation}
To assess the effectiveness of our proposed approach—from pairwise interaction reasoning to full-region social group detection—we conducted two complementary evaluations: one at the relationship level and one at the group region level.

\paragraph{Pairwise-relationships}
We first evaluated the performance of the fine-tuned VLM on the task of predicting whether two individuals in a street-view image belong to the same social group. For this purpose, we used the test split of the fine-tuning dataset, which was additionally revised to ensure a high quality of labels. For each of the answer options (`Yes' and `No'), we compute common binary classification metrics: $F_1$, precision, and recall. We compare our results to the zero-shot performance of available open-source and commercial VLMs as a baseline.

\paragraph{Social Group Region Detection}
To evaluate the quality of group region detections produced by our pipeline, we adopt the ELSA benchmark \cite{hosseini2024elsaevaluatinglocalizationsocial}, as it provides the most relevant content—urban street-view imagery annotated with social group bounding boxes across a range of activities and group sizes.

We evaluate the predicted group region bounding boxes using the mean Intersection over Union (mIoU) to measure spatial alignment with annotated group regions. In addition, we report precision, recall, and $F_1$ score, using an IoU threshold of 0.5, consistent with common object detection evaluation practices.

\subsection{Implementation Details}

We tested several fine-tuning configurations, including LoRA adapters with rank 64 and 128, and full fine-tuning. For fine-tuning, we used 8 NVIDIA A100 GPUs and repeated all experiments twice. Each model was trained for 5 epochs and the best-performing one is selected based on the validation set performance.

\begin{figure*}[!t]
\centering
\includegraphics[width=1.0\textwidth]{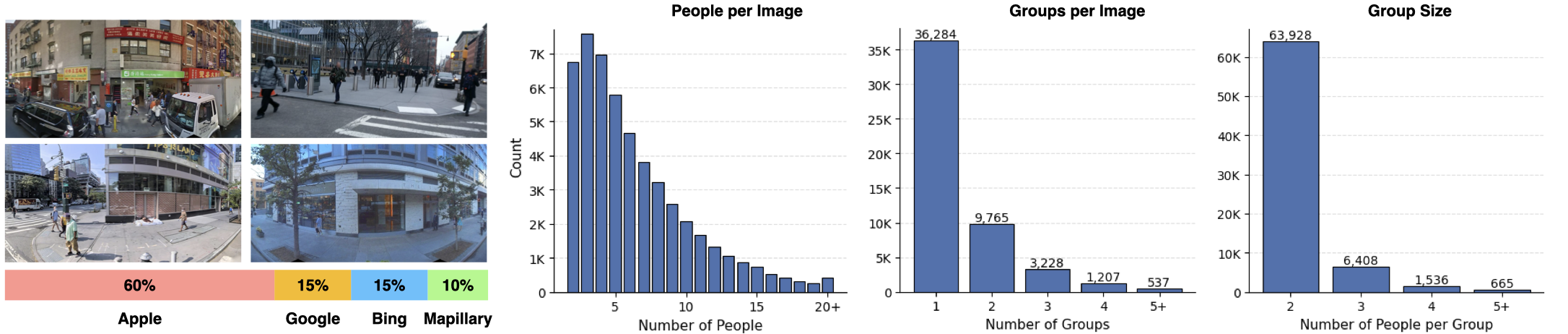}
\caption{Descriptive statistics pertaining to our Social Group Region dataset.}
\label{fig:dataset}
\end{figure*}

\section{Results}
\subsection{Fine-tuning}
To determine the best fine-tuning setup, we conducted a set of experiments on different subsets of the dataset and different preprocessing options (see selected experiments in Table~\ref{tab:fine_tuning_results}). The results illustrate that fine-tuning confers substantial gains over the zero-shot performance. While increasing the dataset size from 29K to 66K or 76K allowed for improving the model's effectiveness and reducing overfitting, data quality and dataset balance are crucial for achieving superior and more robust results. 

Our experiments with full parameter fine-tuning showed promising results. However, given the relatively small training dataset size and resource limitations, they are currently excluded from the scope of this study. 

\subsection{Social Group Region Detection}

To measure the real-world performance of our full pipeline, we evaluate detected social group regions and compare our results to the zero-shot performance of state-of-the-art open source and commercial VLMs (see Table \ref{tab:zero_shot_region_detection}). Our approach substantially outperforms the baselines and illustrates the current inability of all examined models to accurately interpret and localize interpersonal relationships, showing that they remain  inadequate for this challenging task.

\begin{table}[h]
\centering
\begin{tabular}{lcccc}
\textbf{Model} & \textbf{mIoU} & $\boldsymbol{F_1}$ & \textbf{Precision} & \textbf{Recall}\\
\midrule
MINGLE & 0.64 & 0.60 & 0.75 & 0.61 \\
GPT-4o & 0.01 & 0.01 & 0.19 & 0.01 \\
Claude-Sonnet & 0.02 & 0.02 & 0.02 & 0.02 \\
Claude-Opus & 0.01 & 0.02 & 0.20 & 0.02 \\
LLama3.2-Vision & 0.00 & 0.00 & 0.91 & 0.00 \\
Qwen2-VL (7B) & 0.00 & 0.00 & 0.88 & 0.00 \\
\end{tabular}
\caption{Currently available VLMs are unable to reliably detect semantically complex regions.}
\label{tab:zero_shot_region_detection}
\end{table}

\subsection{Distance and Depth Filtering Effects}

Using thresholds for depth and distance allows us to exclude pairs of individuals that are highly unlikely to belong to the same social group, thus decreasing the computational cost and improving the overall effectiveness by reducing false positives. 

\begin{table}[h]
\centering
\begin{tabular}{lccccccc}
\textbf{Filtering} & $\boldsymbol{\tau_d}$ & $\boldsymbol{\tau_z}$ & \textbf{mIoU} & $\boldsymbol{F_1}$ & \textbf{Pairs} \\
\midrule
No Filtering & 255 & 1 & 0.635 & 0.604 & 1071 \\
Optimal Depth & 80 & 1 & 0.634 & 0.603 & 1070\\
Optimal Distance & 255 & 0.4 & 0.636 & 0.608 & 1068 \\
Combined & 80 & 0.4 & 0.634 & 0.607 & 1068 \\
\end{tabular}
\caption{Summary of depth and distance filtering effects on the social group region detection.}
\label{tab:threshold_analysis}
\end{table}

We ran a sweep over all (1) distance values [0,1] with step size 0.1, and (2) depth values [0,255] with step size 20. 

The results demonstrate that the optimal effectiveness is attained at distance values 0.3–0.5 and depth values 80–100. However, the limited size of the benchmark dataset may limit the generalization of this finding, and we recommend re-running a similar analysis on a relevant dataset of interest.

\subsection{Social Group Region Dataset}

We introduce a large-scale street scene dataset comprising 100K annotated images for social group detection research. The dataset combines manual annotations with pipeline-propagated labels to ensure both quality and scale. Images were systematically collected from four major mapping platforms to capture diverse urban environments and imaging conditions: Apple Lookaround (60\%), Google Street View (15\%), Bing Streetside (15\%), and Mapillary (10\%).

To optimize social interaction visibility, we extract lateral viewpoints (left and right) from panoramic imagery, as these perspectives provide superior coverage of pedestrian interactions compared to frontal or rear views. Each image contains a minimum of two human detections, resulting in 588,430 total human instances across the dataset. Ground truth annotations include 72,537 social group regions, with detailed statistics on people-per-image, groups-per-image, and group composition distributions presented in Figure \ref{fig:dataset}.

This dataset addresses the critical gap in training data for semantically complex region detection, providing the first large-scale benchmark specifically designed for social interaction analysis in urban environments from street-view imagery. The combination of diverse data sources, viewpoint selection, and comprehensive annotation coverage makes this dataset suitable for both model training and systematic evaluation of social group detection algorithms.

\section{Conclusion}
In this work, we address the previously unsolved urban analysis challenge: social group region detection. Our extensive zero-shot evaluation confirms that existing models fail to reliably localize social groups, highlighting the semantic complexity of the task. To tackle this, we have presented MINGLE: a modular three-stage VLM-based pipeline, which reframes the problem from direct region-level detection to pairwise social affiliation classification. 

To encourage future work, we release a large-scale street-view dataset annotated with granular information about social groups created through a combination of manual and pipeline-based labeling.

\section{Acknowledgments}
The project was supported by the Designing for Sustainability research program, a joint initiative of the Morningside Academy for Design at MIT and the Hasso Plattner Institute, generously funded by the Hasso Plattner Foundation.

This work used the Jetstream2 GPU and NCSA Delta GPU resources through allocation CIS250088 from the Advanced Cyberinfrastructure Coordination Ecosystem: Services \& Support (ACCESS) program, supported by the U.S. National Science Foundation (NSF) under grants \#2138259, \#2138286, \#2138307, \#2137603, and \#2138296.
\bibliography{aaai2026}

\end{document}